\documentclass[sigconf,nonacm]{aamas}

\usepackage{balance} 
\usepackage{listings}

\setcopyright{ifaamas}
\acmConference[AAMAS '26]{Proc.\@ of the 25th International Conference
on Autonomous Agents and Multiagent Systems (AAMAS 2026)}{May 25 -- 29, 2026}
{Paphos, Cyprus}{C.~Amato, L.~Dennis, V.~Mascardi, J.~Thangarajah (eds.)}
\copyrightyear{2026}
\acmYear{2026}
\acmDOI{}
\acmPrice{}
\acmISBN{}

\acmSubmissionID{216}

\title[Can Vibe Coding Beat Graduate CS Students?]{Can Vibe Coding Beat Graduate CS Students? An LLM vs. Human Coding Tournament on Market-driven Strategic Planning}

\author{Panayiotis Danassis}
\affiliation{
  \institution{University of Southampton}
  \city{Southampton}
  \country{United Kingdom}}
\email{p.danassis@soton.ac.uk}
\thanks{Part of this work was conducted while P.D. was with Telenor Research.\\ Code available at: \benchmarklink{}}

\author{Naman Goel}
\affiliation{
  \institution{University of Oxford and Alan Turing Institute}
  \city{Oxford}
  \country{United Kingdom}}
\email{naman.goel@alumni.epfl.ch}

\begin{abstract}
    The rapid proliferation of Large Language Models (LLMs) has revolutionized AI-assisted code generation. 
    This rapid development of LLMs has outpaced our ability to properly benchmark them. Prevailing benchmarks emphasize unit-test pass rates and syntactic correctness. Such metrics understate the difficulty of many real-world problems that require \emph{planning, optimization, and strategic interaction}. We introduce a multi-agent reasoning-driven benchmark based on a real-world logistics optimization problem (Auction, Pickup, and Delivery Problem) that couples competitive auctions with capacity-constrained routing. The benchmark requires building agents that can (i) bid strategically under uncertainty and (ii) optimize planners that deliver tasks while maximizing profit. We evaluate \emph{40 LLM-coded} agents (by a wide range of state-of-the-art LLMs under multiple prompting methodologies, including vibe coding) against 17 \emph{human-coded} agents developed \emph{before the advent of LLMs}. Our results over 12 double all-play-all tournaments and \emph{$\sim 40$k matches} demonstrate (i) a clear superiority of human(graduate students)-coded agents: the \emph{top 5 spots are consistently won by human-coded agents}, (ii) \emph{the majority of LLM-coded agents (33 out of 40) are beaten by very simple baselines}, and (iii) given the best human solution as an input and prompted to improve upon, the best performing LLM makes the solution significantly worse instead of improving it. Our results highlight a gap in LLMs' ability to produce code \emph{that works competitively in the real-world}, and motivate new evaluations that emphasize reasoning-driven code synthesis in real-world scenarios.
\end{abstract}

\keywords{Large Language Models (LLMs), Code generation, Vibe Coding, Benchmarks, Human Evaluation}

\usepackage{url}
\usepackage{graphicx}
\usepackage{amsmath}
\usepackage{amsthm}
\usepackage{dsfont}
\usepackage{subcaption}
\usepackage{color}
\usepackage{enumerate}
\usepackage{cases}
\usepackage{multicol}
\usepackage{multirow}
\usepackage{enumitem}
\usepackage{bm}
\usepackage{import}
\usepackage{lscape}
\usepackage{xcolor}
\usepackage{float}
\usepackage{booktabs}       
\usepackage{caption}
\usepackage{wrapfig}

\definecolor{BrightBlue}{RGB}{65, 145, 225}
\newcommand{\pan}[1]{{\color{orange}[Panayiotis: #1]}}

\newcommand{\course}[0]{Intelligent Agents course}
\newcommand{\lab}[0]{Artificial Intelligence Laboratory at EPFL}
\newcommand{\uni}[0]{EPFL}
\newcommand{\platform}[0]{Logist platform}
\newcommand{\benchmarklink}[0]{\url{https://panayiotisd.github.io/apdp_bench/}}

\makeatletter
\newcommand\footnoteref[1]{\protected@xdef\@thefnmark{\ref{#1}}\@footnotemark}
\makeatother

\begin{document}


\pagestyle{fancy}
\fancyhead{}


\maketitle 


\section{Introduction} \label{sec: Introduction}

\begin{figure}
    \centering
    \includegraphics[width=0.95\linewidth]{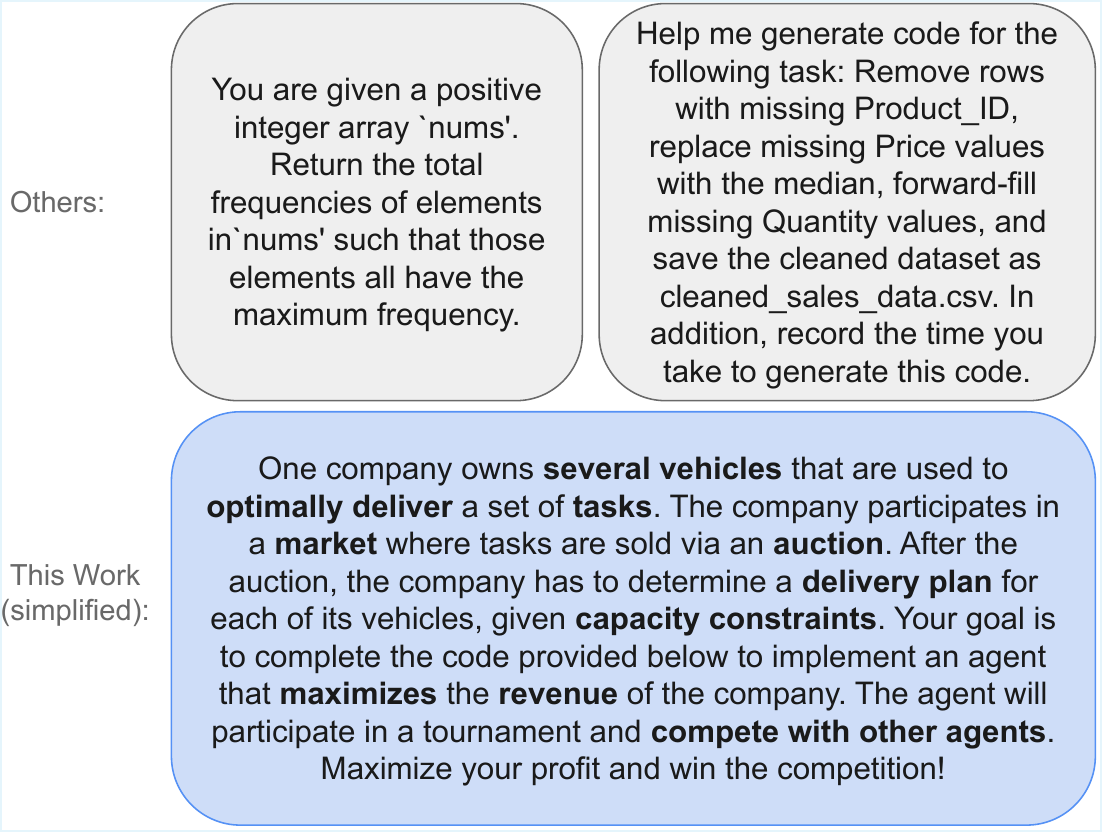}
    \caption{Traditional benchmarks (top) focus on problems with clearly defined correct or incorrect solutions, typically verified through unit tests. In contrast, our benchmark (bottom) involves complex tasks such as planning, constraint optimization, modeling competitors, competitive strategy design, and advanced algorithm development — challenges that remain highly non-trivial even for experienced software engineers. Top from~\cite{jainlivecodebench} (left) and~\cite{huynh2025large} (right).}
    \label{figure: prompt Comparison}
    \Description{Prompt Comparison}
\end{figure}

Large Language Models (LLMs) have demonstrated an impressive ability to generate executable code, free of syntax errors~\cite{achiam2023gpt,anthropicreport,gemini25report}. Software engineers increasingly use LLMs for code generation, bug detection, code refactoring, and more~\cite{google_DORA_report_2025,googleinternalcodeuse,cursorusers}. `Vibe-coding' has empowered users of all technical backgrounds to turn their ideas into code in seconds~\cite{martin_vibe_2025}.\footnote{The term `vibe-coding' colloquially refers to a software development approach that uses LLMs (general-purpose or specialized for coding) to generate code from natural language prompts.} As a result, there is significant interest in evaluating and improving the coding capabilities of LLMs. Recent literature has proposed a plethora of benchmarks evaluating various aspects of code generation such as functional correctness, reliability, robustness, execution time, code security, etc.~\cite{assesing10, huynh2025large, zhong2024can}. Most existing benchmarks are composed of problems whose solutions can be easily verified by running unit tests (e.g., see examples in Figure~\ref{figure: prompt Comparison} (top)). Achieving state-of-art performance on these benchmarks is indeed an impressive milestone for AI. Yet, if we look \emph{beyond autocomplete}, real-world software development is far more complex. We argue that it is time to expand the frontier in code generation even further by asking:
\begin{quote}
	\emph{Does high performance on existing coding benchmarks translate to the ability to solve real-world coding problems, ones requiring multi-agent strategic interaction, planning, optimization, and advanced algorithm design that can be highly non-trivial even for experienced human software engineers?}
\end{quote}

We introduce a real-world optimization benchmark for LLM code generation. The proposed benchmark requires LLMs to complete code that optimizes the operations of a logistics company, specifically the pickup and delivery of parcels (see Figure \ref{figure: problem description}), with the ultimate goal of maximizing profit. The LLMs need to implement an agent that competes against other agents, optimally bids for tasks (tasks are assigned via a reverse first-price sealed-bid auction), and optimally plans for the pickup and delivery of the won tasks. As such, the problem incorporates challenges from the domains of \emph{multi-agent systems (MAS)}, \emph{auctions}, and \emph{constraint optimization}.

Devising a suitable benchmark is one part of the research challenge addressed in this work; the second relates to evaluation metrics. The Auction, Pickup, and Delivery Problem (APDP) is an open ended problem, one that does not admit a closed-form solution (due to real-time constraints for bidding and planning, bounded rationality, etc.). Thus, pass/fail designations of the generated code are not feasible, nor can we estimate how close we are to the optimal solution. Instead, we compare against a wide range of human-coded agents, developed \emph{before the advent of LLMs} (i.e., without any AI-assisted programming). The APDP was given as an assignment in the postgraduate \course{} at \uni{}. Students had two to three weeks to develop an agent, which then competed in a single-elimination tournament for extra course credits. We selected 12 student agents from the class of 2020 and 5 baseline agents developed by members of the \lab{}.

\emph{Our work demonstrates that while state-of-the-art LLMs can generate code that runs (i.e., free of syntax errors), the generated solution is \emph{not} competitive to human-designed solutions on dimensions such as strategic planning, optimization, or multi-agent competition. Thus, this work brings to the forefront this new frontier in code generation, and aims to facilitate the development of benchmarks, datasets, and open-source baselines that stress reasoning-driven code synthesis.}

\subsection*{Summary of our Contributions} \label{sec: Our Contributions}

\begin{figure}[t]
    \centering
    \includegraphics[width=\linewidth]{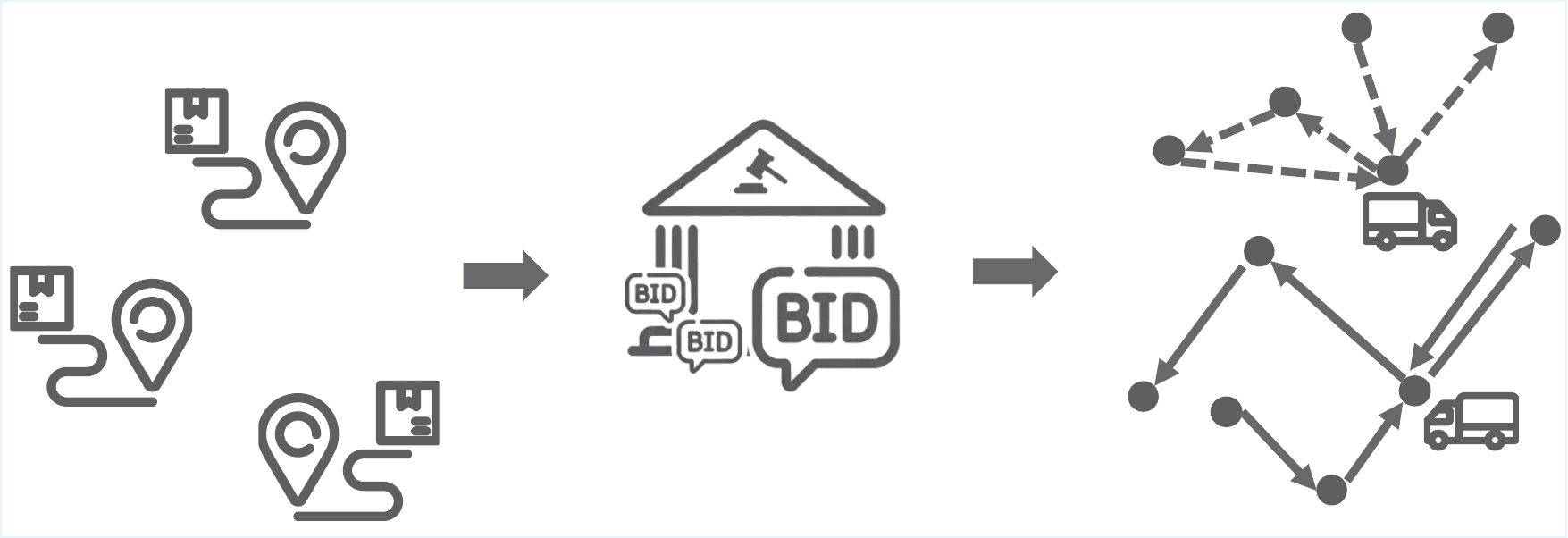}
    \caption{The Auction, Pickup, and Delivery Problem (APDP). Task: Logistic operations optimization. Goal: Maximize profit for a transportation company.\\Multiple transportation companies (agents) compete in a market. Each company owns several vehicles that deliver tasks (e.g., parcels) in a given network. Tasks are sold via a reverse first-price sealed-bid auction, i.e., a company's bid corresponds to the amount of money they want to be paid to deliver the task. Higher bids mean more revenue, but bidding too high may result in not getting the auctioned task. A competitive bid depends on (i) the marginal cost of adding the auctioned task to the partial delivery plan, given the already won tasks, (ii) the marginal cost of the opponent, and (iii) other strategic decisions like incurring a loss (bid below your marginal cost) at the beginning in order to reduce the cost of future tasks (better positioning in the market).\\After the auction is complete, the company has to determine a plan for its vehicles such that all tasks won by the company are delivered and the total revenue of the company is maximized. The plan is a sequence of pickup and delivery actions, such that vehicle capacity constraints are satisfied. The total revenue of the company is defined as the sum of rewards (won bids paid out by the auction house) minus delivery cost (kilometers driven times cost per kilometer). It is, thus, necessary to bid optimally and compute efficient delivery plans.}
    \label{figure: problem description}
    \Description{High Level Problem Description.}
\end{figure}

\begin{enumerate}
    \item \textbf{We provide a novel, reasoning-driven benchmark for automatic code generation.} The Auction, Pickup, and Delivery Problem (Figure~\ref{figure: problem description}) in the proposed benchmark combines challenges from the domains of competitive multi-agent systems, auctions, and constraint optimization and requires capabilities to design and code advanced planning and optimization algorithms. The APDP benchmark will be open-sourced.
    \item This is the \emph{first} work, to the best of our knowledge, that answers two crucial questions: \textbf{How well does LLM code generation and vibe coding perform beyond autocomplete, in real-world planning settings?} \emph{Are Large Language Models graduate-level coders?} We evaluated a range of state-of-the-art LLMs (both paid and free) from various companies, ones that most people would have access to (GPT-5 Thinking, Gemini 2.5 Pro, Claude Opus 4.1, DeepThink R1), against 17 human-coded agents, developed \emph{before the advent of LLMs}, including 12 agents developed by \emph{students}. 
    \item Our results over 12 double all-play-all tournaments and almost \emph{40k matches} (Table~\ref{tb: results}) demonstrate a clear superiority of student-coded agents. (i) The \textbf{top 5 spots are consistently won by student agents}, and (ii) \textbf{the majority of LLM agents (33 out of 40) are beaten by very simple baseline agents} (such as the expected cost fixed bid). Perhaps surprisingly, we also find that \textbf{LLMs degrade the performance of top human-coded solutions}. When the best performing LLM was given as input the winning student solution and asked to improve it, the resulting agent performed \emph{significantly worse}, dropping to 10th place.
    
\end{enumerate}

\section{Related Work \& Discussion} \label{sec: Related Work and Discussion}




There is much literature on evaluating the coding capabilities of LLMs. Claims about LLM capabilities range from LLMs being excellent coding assistants (e.g., by autocompleting lines of code in an IDE)~\cite{cui2025effects,yeverechyahu2024impact} and LLMs being better than humans in competitive coding~\cite{brownchart,li2022competition} to LLMs being able to self-improve their own code~\cite{novikovalphaevolve}. Fully automating software engineering is also considered by many as a step towards ``AGI (Artificial General Intelligence)''~\cite{Amodeipodcast}. 

There exist several evaluation methodologies for coding LLMs. One way to measure the utility of LLM-produced code is to measure the impact on developers' productivity~\cite{becker2025measuring}. A related method is to evaluate the quality of code based on subjective ratings or acceptance rate of code suggestions by humans~\cite{lmarena,mozannarrealhumaneval,googleinternalcodeuse}. While capabilities are continuously evolving with new models being released, prior work have identified various strengths and weaknesses of different models ~\cite{austin2021program,huynh2025large,liu2024exploring}. For example,~\cite{becker2025measuring} found that while developers thought they were 20\% faster with AI tools, they were actually 19\% slower when they had access to AI than when they didn't. Their results (and disagreement with previous studies~\cite{cui2025effects,yeverechyahu2024impact}) raised important questions, e.g., what are the right metrics to evaluate the utility of LLMs in coding tasks, in what ways are coding-LLMs actually useful and where do they still lack capability?

While insightful, such user-centric evaluation methods are also costly and results are difficult to reproduce. 
Therefore, it is more common for evaluation methods to focus on static benchmarks. Existing benchmarks primarily evaluate two things: functional correctness~\cite{chen2021evaluating} (with metrics such as pass@k~\cite{chen2021evaluating}, pass-ratio@n~\cite{yeo2024framework}, CodeBLEU~\cite{ren2020codebleu}, etc.) and efficiency (execution time)~\cite{qiuefficient,du2024mercury} of the generated code. Other criteria such as code security, semantic correctness, code interpretability have also drawn attention~\cite{assesing10, huynh2025large, zhong2024can}. Some representative and most commonly used~\cite{achiam2023gpt,anthropicreport,gemini25report} benchmarks are: HumanEval~\cite{chen2021evaluating}, APPS~\cite{hendrycks2measuring}, MBBP~\cite{austin2021program}, BigCodeBench\cite{zhuobigcodebench}, LiveCodeBench~\cite{jainlivecodebench, whitelivebench}, SWEBench and its variants~\cite{jimenezswe,swebenchv,zhang2025swe, swelong, miserendino2025swe}, Aider Polyglot\cite{aider}, WebDev Arena~\cite{chiang2024chatbot,lmarena}, etc.
We direct the interested readers to~\cite{dong2025survey} for a survey of various coding benchmarks. 
In addition, benchmarks that contain questions, unit tests, and human data from platforms such as Codeforces\footnote{\url{https://codeforces.com/blog/entry/68288}} are often used to compare LLM performance with human coders~\cite{brownchart,li2022competition}.
While all these benchmarks offer a scalable quantitative way to evaluate models (and contain challenging coding problems), they come with a different set of limitations: data contamination (where models train on test data)~\cite{roberts2023cutoff}, limited scope that doesn't reflect real-world and open-ended tasks, lack of adaptability and creative testing, etc. 





Therefore, there is a growing interest in the community to find novel and more robust ways to test the models. For example, \cite{grosnit2025kolbbasedexperientiallearninggeneralist} design and evaluate a code generation system for data science programming problems in Kaggle competitions. While data science solutions (e.g., ML classification models) are evaluated on hidden test sets, they are also different from other coding competition problems in the sense that (i) they are often closer to real-world scenarios and (ii) evaluation metrics are usually non-binary (i.e., even a ``correct'' solution can be improved through, e.g., model selection, feature engineering, hyper-parameter tuning, etc.). 

When it comes to real-world competitions, in July 2025, at the ``AtCoder World Tour Finals 2025 Heuristic'', an OpenAI system competed against human coders and finished second. While there is not much information available about the OpenAI system used in the competition, it was reportedly a custom system (unlike ChatGPT or models available to the public through APIs)~\cite{guardianopenai}. The task was to write code to guide a fleet of robots across a grid, placing barriers to minimize the number of moves~\cite{atcoderproblem}. This was an example of a complex optimization problem and required skills beyond code synthesis and knowledge of data structures. The solutions were scored in a competitive way relative to one another. However, there was no direct competition between different opponents within the problem environment.

Our work takes a step further in this direction by increasing the difficulty of evaluation. 
More specifically, we pick a novel complex domain that : (i) involves complex reasoning in a competitive environment to design an end-to-end solution by combining a number of components (data structures, optimization algorithms, auction mechanisms, modeling the opponents, strategic multi-agent interactions, etc.), (ii) involves generating code for heuristics which can be evaluated based on continuous metrics of quality to compare with other competitors, 
and (iii) has human coding data available (from pre-LLM era) to compare performance.



\section{The Auction, Pickup, and Delivery Problem (APDP)} \label{sec: The Pickup and Delivery Problem}

The Pickup and Delivery Problem (PDP) is part of a broad and significant class of combinatorial optimization problems central to logistics, transportation, and supply chain management~\cite{berbeglia2007static,CAI2023126631} with \emph{large practical value}, as they lie in the \emph{core} of companies like Amazon, DLH, FedEx, etc.\footnote{PDPs are ubiquitous in real-life outside the described package delivery scenario, in domains such as ride-pooling (dial-a-ride)~\cite{Danassis2022}, meal delivery routing~\cite{reyes2018meal}, supply-chain management for manufacturing companies such as Huawei and Tesla~\cite{CAI2023126631}.} Traditionally, these problems set to address the challenge of designing optimal routes for a fleet of vehicles to serve a set of transportation requests (tasks), with the goal of minimizing operational costs while adhering to a complex set of real-world constraints. For more details on the PDP problem and its variants, we refer the interested reader to~\cite{berbeglia2007static,CAI2023126631}.

Our APDP is a variant of the PDP in which tasks are auctioned in a reverse first-price sealed-bid auction. See Figure \ref{figure: problem description} for a high level problem description. Transportation companies managed by agents (sellers of services) compete to obtain business from the buyer (task auctioneer) and prices will typically decrease as the companies underbid each other. The APDP has two stages: (i) In the first stage (auction), each task is auctioned one by one. The auction house publishes the details of the task, i.e., the pick-up city, delivery city, and weight. Agents then submit bids, representing the payment that the agent requests for the delivery of the task. Lowest bid wins. Each task is evaluated by agents \emph{locally}, with \emph{no knowledge of future tasks}\footnote{Agents have access to the task distribution, and they may simulate synthetic future tasks to account for future supply/demand in their planning.}. Agents need to consider the already won tasks (and potential future tasks) to form bundles. (ii) In the second stage (vehicle routing), each agent has to solve a static PDP problem, efficiently scheduling a fleet of vehicles of \emph{different characteristics} (starting location, capacity, cost per kilometer) to pickup and deliver all won tasks. The overarching goal for each agent is to maximize the profit for the company.

\subsection{Problem Formulation} \label{sec: Problem Formulation}

\begin{figure}[t]
    \centering
    \includegraphics[width=1\linewidth, clip, trim={1em 0em 0em 0em}]{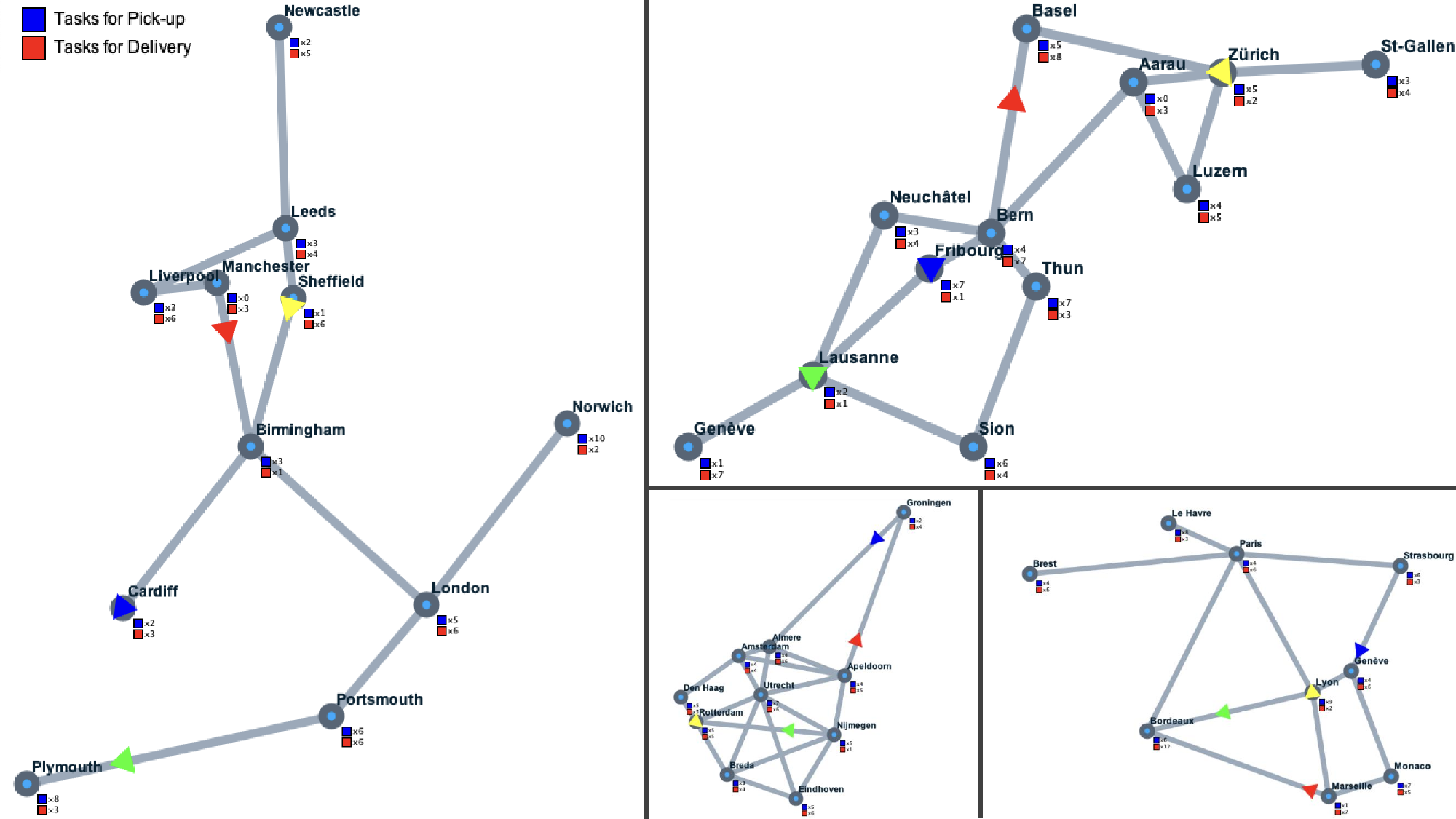}
    \caption{Network Topologies. From top to bottom and left to right we have: Great Britain, Switzerland, the Netherlands, and France. Colored triangles represent vehicles.}
    \label{figure: maps}
    \Description{Network Topologies.}
\end{figure}

The APDP involves the transportation of tasks (e.g., packages) from an origin location to a corresponding destination in an undirected graph $G(V, E)$ (see Figure \ref{figure: maps}). Let $T = \{\tau_1, \tau_2, . . . \tau_{N_T} \}$ denote the tasks to be auctioned, each with a source and destination in $V$, and weight $w_{\tau_i}$. The tasks are drawn from a distribution $\Delta(V \times V \times \mathbb{R}_{>0})$, known to the agents. We assume we already know the shortest path between any two cities. Each agent $i$ manages a company $C_i$ operating a fleet of vehicles $V_i = \{v_{i,1}, v_{i,2}, . . . v_{i, N_V} \}$. For simplicity, we assume $N_V = 2$ (though the coded solutions can work for $N>2$). Each vehicle is characterized by its starting location, capacity, and cost per km. I.e., the agents manage a \emph{heterogeneous} set of companies, each operating a \emph{heterogeneous} fleet of vehicles.

At the start of the game, a central auctioneer auctions task sequentially, in a reverse first-price sealed-bid auction. Each participant then has $t_{bid}$ seconds to submit their bid. After each task is auctioned, the agents can observe their opponents' bids. The number of tasks to be auctioned is \emph{not known} to the participants. At the end of the auction, each participant must submit a valid solution, that satisfy certain constraints (described below). Each agent has $t_{plan}$ seconds to plan. The winner is the agent that maximizes the company's profit ($P_i$), given by $\sum_{\tau \in \text{won}(i)} b_{i,\tau} - c_{i, \tau}$, where $\sum_{\tau \in \text{won}(i)} b_{i,\tau}$ is the total revenue, calculated by summing the agent's winning bids ($b_{i,\tau}$) for all tasks won by agent $i$, and $\sum_{\tau \in \text{won}(i)} c_{i, \tau}$ is the total transportation cost (distance driven $\times$ cost per kilometer for each vehicle) for delivering all tasks in $\text{won}(i)$ according to $i$'s solution.

\paragraph{\textbf{Core Constraints}} A valid solution must satisfy a set of constraints that model operational rules. Specifically: (i) \emph{Capacity:} The cumulative task load at any given time on a vehicle can not exceed its maximum capacity. (ii) \emph{Delivery:} All won tasks must be picked up and delivered. (iii) \emph{Pairing:} A task pickup and its corresponding delivery must be performed by the same vehicle. (iv) \emph{Precedence:} The pickup must precede the delivery for all tasks.

\subsection{Challenges in APDP} \label{sec: Challenges}

APDP incorporates challenges from non-cooperative MAS (in stage 1, as they compete against other strategic agents), cooperative MAS (in stage 2, as they collaborative manage a fleet of vehicles), auctions under uncertainty (as they do not know exact valuations or future bundles), and constraint optimization.

The Pickup and Delivery Problem (PDP) is NP-hard~\cite{CAI2023126631}. The implication of NP-hardness is that there is no known algorithm that can find a provably optimal solution in a time that grows polynomially with the size of the problem instance (e.g., the number of tasks). This inherent computational complexity creates a fundamental trade-off between solution quality (optimality) and computational time (scalability). Thus, an important characteristic of our proposed benchmark is the ability of an agent to find not the single best solution, but one that \emph{effectively navigates the quality-versus-time trade-off} to meet the demands of practical applications.

In APDP, agents additionally face a \emph{series of interconnected auctions}. An agent's action in one auction, directly affects its state and, thus, strategy space in all subsequent auctions, introducing complex dynamics and strategic thinking.\footnote{There is a gap in theoretical properties of truthful mechanisms in dynamic auctions~\cite{los2022strategic}.} 

Some of the major challenges agents face in APDP are as follows:

\begin{itemize}
	\item Vehicles and their capacity are finite. Committing to tasks changes the profitability of future ones. A rational agent must consider both the \emph{marginal cost} (the additional cost incurred to service a new task), and the \emph{opportunity cost} (the expected value of the best alternative task that is foregone, i.e., potential loss of profit) of its actions.
	\item The cost of serving a set of tasks is typically \emph{subadditive}; i.e., the cost of serving tasks A and B together may be less than the cost of serving A plus the cost of serving B. This is due to spatial proximity and the fact that we can combine multiple pickups (economies of scope). Thus, an agent might potentially benefit from strategically underbidding 
    to acquire tasks along desirable routes that could potentially substantially decrease the future marginal cost, and thus allowing to recover the lost profit in the future by undercutting the competition.
	\item Due to the aforementioned synergies between tasks, an agent might realize a significant profit by considering \emph{future tasks} (agents have access to the task distribution) and accounting for expected future combinations (similar to bidding for bundles in combinatorial auctions). This is itself a complex optimization problem as it requires to solve another PDP (NP-hard) for each potential bundle.
	\item When multiple self-interested agents interact, their behavior is governed not only by their individual cost estimates but also by their \emph{strategic considerations of their opponents' actions}. For example, an agent might have an incentive to bid a value different from their true cost to secure a profit, or to mislead the opponent's beliefs about its competitors (opponents' bids are observable). An agent may also strategically lose an auction to manipulate the state of competitors in future auctions. In general, heterogeneity in bounded rationality among agents gives rise to many possible strategies.
	\item Estimating the value of a task must respect \emph{real-time} constraints, as agents have limited time to bid and plan.
\end{itemize}

\emph{These challenges imply that one cannot simply employ an off-the-shelf or memorized algorithm as a competitive solution for the APDP, especially when facing such a wide range of potential opponents. These multifaceted challenges -- combinatorial optimization, sequential decision-making under uncertainty, and strategic interactions -- make APDP a challenging benchmark for the next frontier of code generation.}

\section{Methodology} \label{sec: Methodology}

\paragraph{\textbf{Agents}}

We evaluated 57 agents in total: 40 LLM-coded agents and 17 human-coded agents. 

\paragraph{\textbf{Tournament}}

The agents competed in 12 \emph{double all-play-all} tournaments (using 4 network topologies; 3 tournaments with each topology). Agents in a tournament met every other participant in 1v1 matches. In each match, we assigned a company to each agent. Companies have two vehicles with different capacities, cost per kilometer, and starting location. For fairness, we swap the company each agent controls and repeat the match (i.e., each agent plays against the same opponent once controlling company A and once controlling company B). As such, in each tournament we run 3192 matches ($57\times56$, i.e., double all-play-all). In total, we run \emph{38304 matches} ($3192$ matches $\times12$ tournaments). 

\paragraph{\textbf{Topologies and Tasks}}

We run tournaments in 4 different network topologies that correspond to simplified road network abstractions of Switzerland, France, Great Britain, and the Netherlands (see Figure \ref{figure: maps}). In each match we auctioned \emph{50 tasks}, drawn uniformly at random (a task is defined by its source, destination, and weight).

\subsection{Human-coded Agents}

We compare the LLM agents against 17 human-coded agents, \textbf{developed before the advent of LLMs}. These correspond to 12 agents developed by students of the \course{} and 5 baseline agents developed by members of the \lab{}.

\paragraph{\textbf{The Intelligent Agents Course}}

The Auction, Pickup, and Delivery problem (see Section \ref{sec: The Pickup and Delivery Problem}) was given as an assignment in the postgraduate \course{} at \uni{}.
Similarly, members of the \lab{} developed the \platform{} we used in this work for our simulations. All code is implemented in Java, and is available as an open-source benchmark at \benchmarklink{}.

\paragraph{\textbf{Baseline Agents}}

Members of the \lab{} developed 5 \emph{simple} baseline agents (prior to the advent of LLMs):
\begin{enumerate}
    \item \textbf{Naive}: It calculates the distance to the pickup city of the auctioned task from the current city, plus the distance to the destination, and bids distance $\times$ cost per km (with a small added randomness). Then it uses only a single vehicle (disregarding the rest) to sequentially pickup and deliver each won task (i.e., no intelligent batching/reordering, etc.).
    \item \textbf{ExpCostFixedBid}: It generates 10 random synthetic tasks (random pickup/delivery cities and weights), and estimates an average marginal cost for these tasks. Then it uses this average as its bid for every auctioned task.
    \item \textbf{Honest}: It bids (roughly) the true marginal cost of adding the auctioned task to its current fleet schedule. The marginal cost calculation is done by looking across all vehicles and returning the minimum marginal cost of inserting this task given the current assignments.
    \item \textbf{ModelOpponent}: It calculates the marginal cost (calculated like the Honest) for both itself and its opponent, and bids the maximum between the two marginal costs. For the opponents marginal cost calculation, it keeps track of the opponent's won tasks and simulates a shadow fleet using its own fleet specs as a proxy.
    \item \textbf{RiskSeeking}: It uses an exponential schedule that shifts over time from a rough prior (calculated using synthetic tasks like ExpCostFixedBid) to the current marginal cost (calculated like the Honest). Additionally, it keeps track of the opponent's won tasks and using a shadow fleet (like ModelOpponent), and calculates the same blended cost (using the aforementioned exponential schedule). Finally, it bids the maximum between its own and the opponent's blended cost.
\end{enumerate}

\paragraph{\textbf{Student Agents}}

We have chosen 12 student agents developed (in 2-3 weeks time, along with their normal coursework) as part of the \course{} in \emph{2020} (2 years prior to the introduction of ChatGPT). These agents were chosen as follows. At the end of the \course{}, students competed in a \emph{single-elimination tournament}. We chose the top 8 agents from the 2020 tournament. Note that due to the nature of single-elimination tournaments, these top 8 agents are not necessarily the top human-coded agents overall (in fact, some of them fail to beat the Baseline Agents). We also selected 4 additional student agents that performed well (highest number of wins) against the aforedescribed Baseline Agents.

\subsection{LLM-coded Agents}

We employed 4 different state-of-the-art LLMs: OpenAI's \textbf{GPT-5 Thinking}, Google's \textbf{Gemini 2.5 Pro}, Anthropic's \textbf{Claude Opus 4.1}, and DeepSeek's \textbf{DeepThink R1}. We also used 5 different prompting strategies, as explained below. Each LLM was prompted twice with each of the 5 prompting strategies, for a total of \emph{40 LLM-coded agents}.

\paragraph{\textbf{Prompting Strategies}}

We used the following strategies:
\begin{enumerate}
    \item \textbf{Author Prompt \#1 (A1):} Prompt generated by one of the authors, as detailed in the following paragraph.
    \item \textbf{Author Prompt \#2 (A2):} Prompt generated by a different author.
    \item \textbf{Iterative Refinement (IR):} We evaluate the agent generated using prompt A1 in \emph{self-play}, and we provide the scores as input back to the respective LLM (the one that generated the A1 agent), along with the following addition to the prompt: `Would you like to improve anything, or shall we enter the tournament? We need to win!'.\footnote{Note that in our application the feedback signal is weak, compared to iterative refinement in other coding benchmarks. An interesting avenue to investigate in future work is whether asking the LLM to include prints of debug information (as a human software developer would do) would make a difference.}
    \item \textbf{LLM as a Critic (CR):} We feed the code generated by an LLM (using the A1 prompt) to GPT-5 Thinking (the top performing LLM in our test), along with a short problem description, and ask GPT-5 to `provide a list of improvements to make sure that the agent maximizes profit and wins the competition!'. Then we provide the suggested improvements to the original LLM (the one that generated the A1 agent) and ask to implement them.
    \item \textbf{GPT-5 generated (optimized) prompt (GEN):} We gave the A1 prompt to GPT-5 Thinking (the top performing LLM in our test) and asked it to optimize the prompt. Then, we used the resulting prompt for code generation.
\end{enumerate}

The supplement contains the complete prompts. Below we provide a description of one of the prompts as an example.

\paragraph{\textbf{Description of Author Prompt \#1}}

The prompt was written to contain the same information as the students received in the \course{} project. This is because we want to investigate if LLMs are capable of competing with humans (students), when they are provided with same instructions. This would also correspond to a modern-day student using an LLM to complete their project, or people using vibe-coding approaches.

The A1 prompt aggregated information from the project description, the project presentation slides, and relevant pitfalls we discussed with students during the lab sessions over the years the authors were part of the course. The information was cleaned and structured according to known best practices\footnote{\label{fn: Best practices}Author generated prompts (strategies \#1 and \#2) followed best practices, e.g.,~\cite{Anthropic_Prompt_Engineering,Anthropic_Prompting_guidance}.}. In short, the A1 prompt contained the following information:

The LLM is assigned the role of an expert AI/ML coder. The prompt then provides a high level problem description/summary (similar to what is shown in Figure~\ref{figure: problem description}). Then the prompt describes in detail the vehicle planning problem formulated as a constraint satisfaction problem, using \LaTeX ~to describe variables, constraints, and the cost function. The prompt stresses that the LLM does not need to follow this formulation; instead it may follow any alternative approach. It further stresses that the LLM must find the plan that maximizes the revenue of the company. This is followed by the description of the auction rules, and the strategic decisions the LLM should consider. The prompt continues with a list of tasks the LLM needs to complete, the competition rules, and the \emph{template code} that the LLM needs to complete. Finally, at the very end, the prompt summarizes key points and tasks, as follows:

\begin{itemize}
    \item You need to implement an agent that (i) competes against other agents in a tournament, (ii) optimally bids for tasks, and (iii) optimally delivers the won tasks.
    \item Tasks are allocated via a first-price reverse auction. In other words, you are bidding on how much money the auction house will pay you to deliver a task.
    \item You have to deliver all the tasks you win.
    \item The total revenue is defined as the sum of rewards (won bids paid by the auction house) minus delivery cost (km driven times cost per km). As such, bidding optimally and computing an efficient plan are of paramount importance.
    \item You do not have to implement Stochastic Local Search for the delivery planning. Implement the best option you can. Make sure that your agent obeys the time limits.
    \item Most importantly, you need to implement the best possible solution, the one that maximizes profit and wins the competition!
\end{itemize}

\paragraph{\textbf{Debugging}}

We tested all LLMs on both self-play and tournament conditions until all bugs we could identify got resolved. Each LLM was responsible for fixing the bugs in its code (by prompting the LLM with the error info). We observed a \emph{minimal number of syntactic errors}, but a \emph{significant number of semantic bugs}. Notable examples include: (i) Not respecting the time-out limits (each agent has a fixed time for planning and bidding) despite being given the template code that implements this functionality and being explicitly instructed to adhere to the time limits. (ii) Agents not picking up and/or not delivering tasks they have won in the auction, despite being explicitly instructed to do so. We observed that this was due to either ignoring instructions, or due to logic errors (e.g., removing tasks from a full vehicle during re-planning and then omitting to plan for those). (iii) Violating capacity constraints, where the agents would try to pickup a task that the vehicle could not carry.

Another common issue we found (mostly with Gemini, Claude, and DeepSeek, and not so much with GPT) is that quite often the LLM would consistently fail to resolve a bug. For example, an agent would consistently time-out, despite multiple (e.g., $5-15$) cycles of prompting the LLM with the error and receiving the updated version of the code. The only solution we found for such situations (where the LLM repeatedly fails to resolve the exact same bug) is to \emph{re-start from scratch}. Overall, we observed the need for \emph{significant manual effort to achieve bug-free code}. We had to generate substantially more agents to get the 40 bug-free ones we evaluated\footnote{A related observation was made in~\cite{chen2021evaluating}, where authors find that `repeated sampling from the model is a surprisingly effective strategy for producing working solutions to difficult prompts'.}.

\begin{table*}[]
\caption{Table of results across the 12 double all-play-all tournaments (4 network topologies, 3 tournaments with each topology, totaling almost 40k matches). Each agent in a tournament plays 112 matches, thus the upper limit for the Avg \#Wins / Tour and Avg \#Losses / Tour is 112.  SD = Standard deviation. Human-coded agents in bold. The naming scheme for the LLMs is as follows. The first letter in the parenthesis refers to the model: O = OpenAI’s GPT-5 Thinking, G = Google’s Gemini 2.5 Pro, A = Anthropic’s Claude Opus 4.1, and D = DeepSeek’s DeepThink R1. The next two letters refer to the prompting scheme (see Section \ref{sec: Methodology}). The last number refers to the 1st or 2nd agent we generated (we run each LLM/prompt combination twice, see Section \ref{sec: Methodology}).}
\label{tb: results}
\resizebox{\textwidth}{!}{
\begin{tabular}{@{}rccccccc@{}}
\toprule
\textbf{Agent}        & \textbf{Avg \#Wins / Tour} & \textbf{SD \#Wins / Tour} & \textbf{Avg \#Losses / Tour} & \textbf{SD \#Losses / Tour} & \textbf{Total Wins} & \textbf{Total Losses} & \textbf{Winrate} \\ \midrule
\textbf{Student 1}     & 108.167   & 1.193     & 3.833       & 1.193       & 1298        & 46            & 0.9658         \\
\textbf{Student 2}     & 104.917   & 2.539     & 7.083       & 2.539       & 1259        & 85            & 0.9368         \\
\textbf{Student 3}     & 103.917   & 2.466     & 8.083       & 2.466       & 1247        & 97            & 0.9278         \\
\textbf{Student 4}     & 103.25    & 1.815     & 8.75        & 1.815       & 1239        & 105           & 0.9219         \\
\textbf{Student 5}     & 96.5      & 2.908     & 15.5        & 2.908       & 1158        & 186           & 0.8616         \\
LLM(O, IR, 1)          & 95.417    & 2.314     & 16.583      & 2.314       & 1145        & 199           & 0.8519         \\
LLM(O, A2, 1)          & 94.583    & 2.314     & 17.417      & 2.314       & 1135        & 209           & 0.8445         \\
\textbf{Student 6}     & 93.167    & 1.899     & 18.833      & 1.899       & 1118        & 226           & 0.8318         \\
\textbf{Student 7}     & 93.167    & 3.563     & 18.833      & 3.563       & 1118        & 226           & 0.8318         \\
LLM(O, A1, 1)          & 86.083    & 3.029     & 25.917      & 3.029       & 1033        & 311           & 0.7686         \\
LLM(O, GEN, 2)         & 84.083    & 6.947     & 27.917      & 6.947       & 1009        & 335           & 0.7507         \\
LLM(O, CR, 2)          & 83.5      & 4.442     & 28.5        & 4.442       & 1002        & 342           & 0.7455         \\
\textbf{Student 8}     & 83.417    & 4.122     & 28.583      & 4.122       & 1001        & 343           & 0.7448         \\
\textbf{RiskSeeking}   & 82.417    & 3.343     & 29.583      & 3.343       & 989         & 355           & 0.7359         \\
LLM(O, GEN, 1)         & 80.667    & 4.355     & 31.25       & 4.372       & 968         & 375           & 0.7208         \\
\textbf{ModelOpponent} & 80.583    & 3.26      & 31.417      & 3.26        & 967         & 377           & 0.7195         \\
LLM(D, A1, 1)          & 79.417    & 3.965     & 32.583      & 3.965       & 953         & 391           & 0.7091         \\
\textbf{ExpCostFixedBid}      & 77.167    & 4.951     & 34.833      & 4.951       & 926         & 418           & 0.689          \\
LLM(O, IR, 2)          & 73.917    & 3.502     & 38          & 3.618       & 887         & 456           & 0.6605         \\
LLM(O, A1, 2)          & 72.417    & 2.193     & 39.583      & 2.193       & 869         & 475           & 0.6466         \\
LLM(G, A1, 2)          & 68.5      & 3.555     & 43.5        & 3.555       & 822         & 522           & 0.6116         \\
LLM(A, GEN, 2)         & 67.917    & 2.968     & 44.083      & 2.968       & 815         & 529           & 0.6064         \\
LLM(G, IR, 2)          & 65.917    & 2.314     & 46.083      & 2.314       & 791         & 553           & 0.5885         \\
\textbf{Student 9}     & 64.167    & 11.044    & 47.833      & 11.044      & 770         & 574           & 0.5729         \\
LLM(G, A1, 1)          & 64        & 4.243     & 47.917      & 4.316       & 768         & 575           & 0.5719         \\
LLM(G, IR, 1)          & 60.333    & 3.725     & 51.667      & 3.725       & 724         & 620           & 0.5387         \\
LLM(O, A2, 2)          & 59.333    & 4.499     & 52.667      & 4.499       & 712         & 632           & 0.5298         \\
LLM(D, CR, 1)          & 55.083    & 6.694     & 56.833      & 6.59        & 661         & 682           & 0.4922         \\
LLM(G, GEN, 2)         & 53.167    & 3.664     & 58.833      & 3.664       & 638         & 706           & 0.4747         \\
LLM(D, GEN, 2)         & 52.083    & 9.06      & 59.917      & 9.06        & 625         & 719           & 0.465          \\
\textbf{Honest}        & 50.583    & 3.848     & 61.417      & 3.848       & 607         & 737           & 0.4516         \\
\textbf{Student 10}    & 48.833    & 2.98      & 63.167      & 2.98        & 586         & 758           & 0.436          \\
LLM(D, IR, 1)          & 48.583    & 10.211    & 63.417      & 10.211      & 583         & 761           & 0.4338         \\
LLM(A, A1, 1)          & 48        & 4.69      & 64          & 4.69        & 576         & 768           & 0.4286         \\
LLM(G, A2, 1)          & 47.25     & 3.864     & 64.75       & 3.864       & 567         & 777           & 0.4219         \\
LLM(A, CR, 1)          & 43.833    & 4.609     & 68.167      & 4.609       & 526         & 818           & 0.3914         \\
LLM(A, A1, 2)          & 43.75     & 2.05      & 68.25       & 2.05        & 525         & 819           & 0.3906         \\
\textbf{Student 11}    & 42.083    & 5.664     & 69.917      & 5.664       & 505         & 839           & 0.3757         \\
LLM(A, IR, 1)          & 39.5      & 2.541     & 72.5        & 2.541       & 474         & 870           & 0.3527         \\
\textbf{Naive}         & 36.75     & 1.712     & 75.25       & 1.712       & 441         & 903           & 0.3281         \\
\textbf{Student 12}    & 36.333    & 1.775     & 75.667      & 1.775       & 436         & 908           & 0.3244         \\
LLM(D, A2, 1)          & 33.917    & 2.193     & 78.083      & 2.193       & 407         & 937           & 0.3028         \\
LLM(A, GEN, 1)         & 30.167    & 1.749     & 81.833      & 1.749       & 362         & 982           & 0.2693         \\
LLM(D, A2, 2)          & 29.833    & 2.038     & 82.167      & 2.038       & 358         & 986           & 0.2664         \\
LLM(G, A2, 2)          & 27        & 2.256     & 85          & 2.256       & 324         & 1020          & 0.2411         \\
LLM(A, A2, 1)          & 26.333    & 0.985     & 85.667      & 0.985       & 316         & 1028          & 0.2351         \\
LLM(O, CR, 1)          & 25        & 3.411     & 87          & 3.411       & 300         & 1044          & 0.2232         \\
LLM(A, IR, 2)          & 24.333    & 8.542     & 87.667      & 8.542       & 292         & 1052          & 0.2173         \\
LLM(A, A2, 2)          & 24        & 1.809     & 88          & 1.809       & 288         & 1056          & 0.2143         \\
LLM(A, CR, 2)          & 23.333    & 1.557     & 88.667      & 1.557       & 280         & 1064          & 0.2083         \\
LLM(D, GEN, 1)         & 22.5      & 1.784     & 89.5        & 1.784       & 270         & 1074          & 0.2009         \\
LLM(D, A1, 2)          & 13.333    & 1.826     & 98.667      & 1.826       & 160         & 1184          & 0.119          \\
LLM(G, CR, 1)          & 9.5       & 1.087     & 102.5       & 1.087       & 114         & 1230          & 0.0848         \\
LLM(G, GEN, 1)         & 9.167     & 0.937     & 102.833     & 0.937       & 110         & 1234          & 0.0818         \\
LLM(D, IR, 2)          & 7.75      & 0.622     & 104.25      & 0.622       & 93          & 1251          & 0.0692         \\
LLM(G, CR, 2)          & 7.25      & 1.422     & 104.75      & 1.422       & 87          & 1257          & 0.0647         \\
LLM(D, CR, 2)          & 5.667     & 0.985     & 106.333     & 0.985       & 68          & 1276          & 0.0506         \\ \bottomrule
\end{tabular}
}
\end{table*}

\section{Results} \label{sec: Results}

\subsection{Preliminary Results}


Before diving into the results of the tournament, we discuss some insights from a preliminary experiment in which we tested the LLMs' ability to solve simpler variants of the APDP problem. In these variants, the agents do not participate in an auction, and instead solve simpler versions of only the task planning part. These variants are as follows. (i) A simple \emph{Reactive} agent which moves from city to city and if there is a task available to pick up at the current city (tasks drawn from a known distribution) it has to decide whether to accept it and immediately deliver it, or reject it in which case it will be lost and the agent will move on to another city. This variant involves modeling the problem as an MDP and solving it using value iteration. Then the agent selects actions based on a learned state-action table. (ii) A \emph{Deliberative} agent which, unlike the reactive one, it knows in advance the list of tasks that must be delivered. As such, the goal is to construct a plan that guarantees the optimal delivery. It does so by modeling all possible states and using BFS and A* to find the optimal plan. Finally, (iii) a \emph{Centralized} agent, which faces the PDP constraint optimization problem (Section \ref{sec: The Pickup and Delivery Problem}), but there is no auction or multiple companies involved. The problem involves implementing the Stochastic Local Search algorithm. 

\paragraph{\textbf{Reactive}} 

All LLMs solved this test-case correctly on the first try, with minimal syntax errors (that they resolved when provided with the compiler error logs).

\paragraph{\textbf{Deliberative}}

In this test-case, we observed a surprising result: 3 out of the 4 LLMs (all but GPT-5) \textbf{did not implement an admissible heuristic for A*} (the first time), despite being explicitly asked to do so in the prompt (emphasized in multiple points and in the summary of the task at the end of the prompt). Without an admissible heuristic A* overestimates the cost of reaching the goal, and thus prunes optimal paths. It is a fundamental requirement for A* to work, and the first thing any student learns about A*. 

In general, we observed very inconsistent behavior. For example, one of the Claude agents started with an admissible heuristic, but when asked to fix the bugs because the solution was not optimal, it changed the heuristic to an inadmissible one\footnote{We did not specify what is the bug, but instead let the LLM figure it out itself.}. In another example, we used GPT4o as a critic, but Claude did not manage to fix the issues GPT4o raised, and the heuristic was still inadmissible\footnote{Contrary to that, GPT4o was able to update Claude’s code and fix the issues.}.

Even when LLMs were providing admissible heuristics, it is worth considering that not all heuristics are equally good. Some allow for more aggressive pruning, resulting in a significantly faster solution. For example, DeepThink R1 was the only LLM that opted to implement a Minimum Spanning Tree based heuristic, which is significantly tighter that any of the heuristics implemented by the rest of the LLMs, allowing it to gracefully scale to larger problem sizes. As a result, DeepThink R1's agent was \emph{one order of magnitude} faster than both GPT's and Claude's.

The A* algorithm is one of the most established and well known classic algorithms in computer science and artificial intelligence. Thus, it is surprising that state-of-the-art LLMs struggled to implement an admissible heuristic (not to mention their inability to implement a tight heuristic).

\paragraph{\textbf{Centralized}}

Similar to the Deliberative test-case, the LLMs were able to produce syntax-bug-free code, but we often observed suboptimal design decisions. For example, in this test-case the LLMs have to schedule multiple vehicles. We observed situations where the resulting agent would utilize only one vehicle from its fleet, despite being specifically prompted to find the optimal solution, and despite the LLM itself noticing this inefficiency and suggesting to improve it: `Let me know if you want to further improve the solution by [\dots] supporting multiple pickups before delivery'. Finally, we compared the LLM-coded agents to a simple template code,\footnote{A very basic implementation of SLS, without any optimizations, instead designed to be easily read and understood by the students of the \course{}. It starts by assigning all tasks to the largest vehicle, and then generates neighbors by randomly changing the task order inside a vehicle or moving a task to a different vehicle.} and observed a large variation in profit, ranging from 19\% higher to 194\% lower (compared to the template code).

\subsection{Tournament Results}

The observations from the preliminary evaluation indicate that LLMs did not generate expected/competitive code even in simpler variants of the APDP problem (despite the code being largely syntax-bug-free). This underlines the importance of reasoning-driven code evaluation benchmarks that go beyond auto-complete and identify new weaknesses of LLMs. We now demonstrate the implications of these weaknesses on the full APDP problem variant.

Table \ref{tb: results} contains the scores for the 12 double all-play-all tournaments (\emph{$\sim$40k matches}). We report the average number of wins and losses per tournament (bounded by 112 -- the number of matches for each agent per tournament), the standard deviation, the total number of wins and losses across all tournaments, and the win rate.

Our results demonstrate a clear superiority of human-coded agents: (i) The \textbf{top 5 spots are consistently held by student agents}, and (ii) \textbf{the majority of LLM agents (33 out of 40) are beaten by very simple baseline agents} (such as the expected cost fixed bid).

Importantly, we did not debug the student code (while we thoroughly tested/debugged the LLM code, both in self-play and tournament settings, see Section \ref{sec: Methodology}). Every time a student agent crashed, we automatically gave the win to the LLM. A large number of these crashes would be easy to fix (e.g., agents timed-out), thus student agents could potentially \emph{rank even higher}.

For reference, Student 9 crashed on average 15,33 $\pm$ 13,44 times per tournament (184 in total across tournaments), Student 11 6,17 $\pm$ 0,94 times (74), Student 4 5,75 $\pm$ 1,06 times (69), Student 7 3,42 $\pm$ 1,08 times (41), Student 8 2,00 $\pm$ 0,00 times (24), Student 10 2,00 $\pm$ 0,00 times (24), Student 5 0,42 $\pm$ 0,90 times (5), Student 1 0,25 $\pm$ 0,62 times (3), Student 6 0,08 $\pm$ 0,29 times (1).


\subsection{Improving the Tournament Winner?}

It is evident from all the results so far that LLMs underperform compared to student-coded agents in our APDP benchmark. In a further experiment, we asked GPT-5 Pro (the best performing LLM in our benchmark) to improve the code of the winning agent (Student 1). We provided a prompt with the description of the problem (highlighting that the agent has to bid optimally, plan efficient vehicle routes, maximize profit, obey all time constraints, and win the tournament), the code of Student 1, and asked it to implement any improvements it deems necessary to win the tournament. Surprisingly, the resulting agent \textbf{ranked 10th} (just below Student 7), with an average number of wins of 89.667. In other words, \textbf{the LLM's `optimizations' resulted in losing 9 spots in the leaderboard}. 

Many benchmarks are often accompanied with \emph{data contamination concerns} (i.e., the correct answers were already in the training data). On the other hand, this experiment shows that, in our benchmark, \textbf{even when we expose a good solution in-context, the LLM is still unable to utilize it}. This result also raises interesting future research questions about the limits of in-context learning and retrieval-augmented problem solving in complex scenarios.

\section{Implications, Limitations, and Future Work} \label{sec: Implications}

We considered the setting of using LLMs as a tool to generate code (vibe coding), with the aim to evaluate the performance of a representative sample of various LLMs (both paid and free) from various companies, ones that most people would have access to. This is the \emph{first} comparison of LLMs vs. humans in this new frontier in automated code generation.

Our work is \emph{not} concerned with finding the optimal prompting strategy, the best LLM, or to create a leader-board of LLMs. Our conclusions are bounded by a single, albeit rich, domain (logistics PDP). We also do not claim that LLM performance observed in this paper is the optimal performance the best LLM can ever achieve in this setting with any prompt. 
It is clear that some LLMs are better at some tasks than others, and there are even dedicated models, specifically fine-tuned for coding. Just like professional software engineers could presumably be better than the students in this study, and some engineers with more experience would be better than others.
But the \emph{wide range} of employed LLMs and prompting strategies allows us to evaluate, for the first time, \emph{vibe coding} as a means of solving complex optimization problems. The appeal of vibe coding is that it empowers people of all technical backgrounds, not just those who are up to date with the latest prompting recommendations or those that have access to custom models. This is also the reason for not directly reporting the name of each LLM in Table~\ref{tb: results}, but instead including it as secondary information inside a parenthesis, as our aim was to investigate what the state-of-the-art LLMs can do overall compared to humans, instead of creating a leaderboard of LLMs.


Our results highlight important limitations of LLM code generation, most notably their limited reasoning and planning capabilities while generating code (see also~\cite{kambhampati2024can}). Modern LLMs are able to provide syntax-bug-free code that runs, but that is not the benchmark we should be using to measure progress towards advanced general AI. The results call for the development of a suite of next generation code synthesis benchmarks, grounded in real-world settings, 
involving aspects of both multi-agent competition and collaboration. 








\section{Conclusion} \label{sec: Conclusion}

In this work, we investigate whether LLMs' strong reported performance on unit-tests translates to competence in solving real-world software engineering problems, ones requiring planning, optimization, and advanced algorithm design. The undeniably impressive performance of LLMs often gives the impression that vibe coding can spin up software on demand without requiring any technical background by the user. Yet, results from our Auction, Pickup, and Delivery Problem (APDP) benchmark (a multi-agent, strategic-reasoning-driven optimization benchmark), demonstrate a clear superiority of human-coded agents in problems requiring long-horizon planning, opponent reasoning, and strategic optimization. Human-coded agents (developed by graduate students before the advent of LLMs) consistently occupy the top five positions, and the majority of the LLM-coded agents (33/40) lose to simple baselines. These results are a call to explore a new frontier in code synthesis; shifting the goal from \emph{code that compiles} to \emph{code that competes}.

\begin{acks}
    We are grateful to Professor Emeritus Boi Faltings and the members of the Artificial Intelligence Laboratory at EPFL who contributed in the development of the Intelligent Agents course over the years. We also thank the students who contributed their code as baselines. The list of contributors is available at: \benchmarklink{}.
\end{acks}



\balance
\bibliographystyle{ACM-Reference-Format} 
\bibliography{AAMAS_2026_LLM_Coding}


\end{document}